\documentclass[preprint,12pt]{elsarticle}

\usepackage{amssymb}
\usepackage{amsmath}
\usepackage{booktabs}
\usepackage{multirow}
\usepackage{arydshln}
\usepackage{wrapfig}

\journal{Pattern Recognition}

\begin{document}

\begin{frontmatter}



\title{Semantic-E2VID: a Semantic-Enriched Paradigm \\ for Event-to-Video Reconstruction}


\author[1]{Jingqian Wu} 
\author[2]{Yunbo Jia}
\author[2]{Shengpeng Xu}
\author[3]{Boxin Shi}
\author[1]{Edmund Y. Lam}
\affiliation[1]{organization={Department of Electrical and Electronic Engineering, \\ The University of Hong Kong},
            city={Hong Kong},
            country={China}}
\affiliation[2]{organization={School of Artificial Intelligence, \\ Beijing University of Posts and Telecommunications},
            city={Beijing},
            postcode={100876}, 
            country={China}}
\affiliation[3]{organization={State Key Laboratory of Multimedia Information Processing and National Engineering Research Center of Visual Technology, \\ School of Computer Science, Peking University},
            city={Beijing},
            postcode={100871}, 
            country={China}}
\thanks{\textsuperscript{\dag} Corresponding author: elam@eee.hku.hk}

\begin{abstract}
Event cameras provide a promising sensing modality for high-speed and high-dynamic-range vision by asynchronously capturing brightness changes.
A fundamental task in event-based vision is event-to-video (E2V) reconstruction, which aims to recover intensity videos from event streams.
Most existing E2V approaches formulate reconstruction as a temporal--spatial signal recovery problem, relying on temporal aggregation and spatial feature learning to infer intensity frames.
While effective to some extent, this formulation overlooks a critical limitation of event data: due to the change-driven sensing mechanism, event streams are inherently semantically under-determined, lacking object-level structure and contextual information that are essential for faithful reconstruction.
In this work, we revisit E2V from a semantic perspective and argue that effective reconstruction requires going beyond temporal and spatial modeling to explicitly account for missing semantic information.
Based on this insight, we propose \textit{Semantic-E2VID}, a semantic-enriched end-to-end E2V framework that reformulates reconstruction as a process of semantic learning, fusing and decoding.
Our approach first performs semantic abstraction by bridging event representations with semantics extracted from a pretrained Segment Anything Model (SAM), while avoiding modality-induced feature drift.
The learned semantics are then fused into the event latent space in a representation-compatible manner, enabling event features to capture object-level structure and contextual cues.
Furthermore, semantic-aware supervision is introduced to explicitly guide the reconstruction process toward semantically meaningful regions, complementing conventional pixel-level and temporal objectives.
Extensive experiments on six public benchmarks demonstrate that Semantic-E2VID consistently outperforms state-of-the-art E2V methods.
By explicitly modeling and leveraging semantic information, our framework produces sharper reconstructions with clearer object boundaries, stronger semantic consistency, and improved robustness under sparse or ambiguous event conditions.
\end{abstract}

\begin{graphicalabstract}
\centering
\includegraphics[width=\linewidth]{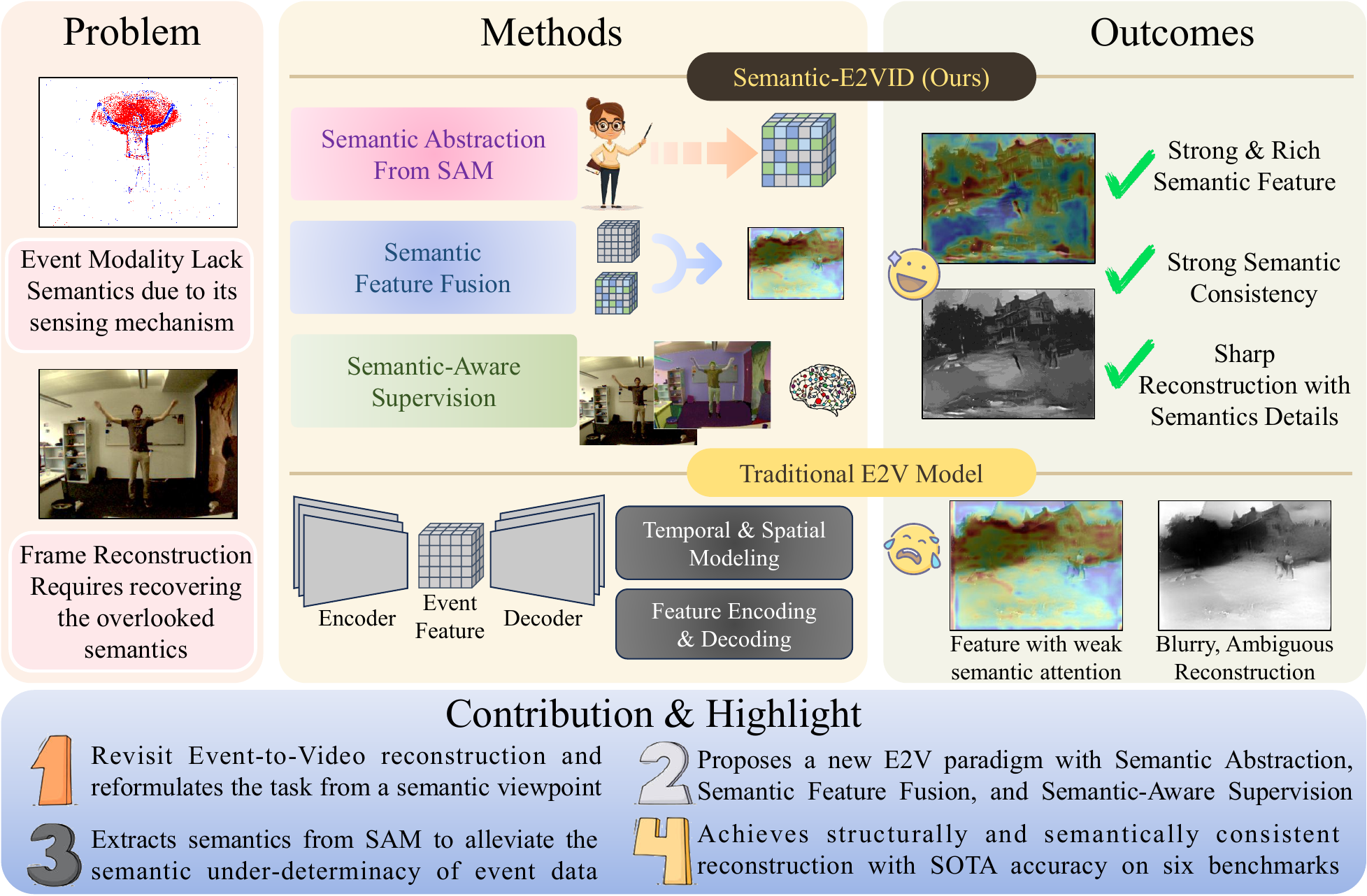}
\end{graphicalabstract}

\begin{highlights}
\item Reformulates event-to-video reconstruction from a semantic viewpoint
\item Proposes a new E2V paradigm with Semantic Abstraction, Semantic Feature Fusion, and Semantic-Aware Supervision
\item Extracts semantic knowledge to alleviate the semantic under-determinacy of event data
\item Achieves structurally and semantically consistent reconstruction with SOTA accuracy on six benchmarks
\end{highlights}

\begin{keyword}
Event Camera \sep Event-based Vision \sep Event-based Video Reconstruction.


\end{keyword}

\end{frontmatter}



\section{Introduction}
\label{sec: intro}

\begin{figure}[h]
	
	\centering
	
	\includegraphics[width=\linewidth,scale=1.0]{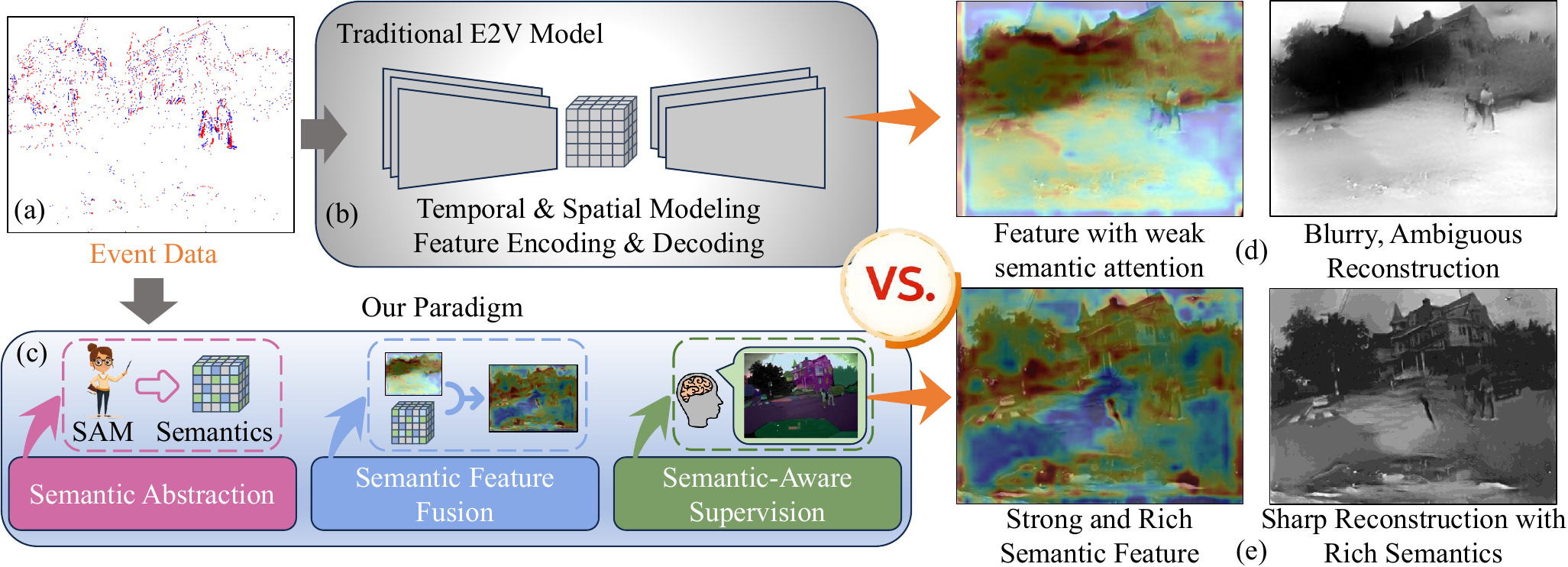}
	
	\caption{
(a) Events capture only sparse brightness changes and inherently lack rich semantic information.
(b) Traditional E2V models primarily rely on temporal and spatial modeling through feature encoding and decoding, which leads to features with weak semantic awareness.
(d) As a result, reconstructed frames often exhibit a blurry appearance and ambiguous object structures.
(c) In contrast, our paradigm explicitly incorporates semantic abstraction, semantic feature fusion, and semantic-aware supervision.
Semantic knowledge is extracted from the pretrained SAM model and integrated into the latent space to form event deep features with rich semantics.
(e) This paradigm enables the reconstruction of sharp frames with stronger semantic consistency, clearer object boundaries, and more faithful structural details.
}

	\label{fig:teaser}
	
\end{figure}

Inspired by the human retina, neuromorphic event sensors are designed to overcome the inherent limitations of conventional frame-based cameras by outputting a novel sensing modality known as events.
Unlike standard cameras that capture image intensities at fixed frame rates, event cameras operate asynchronously and respond only to changes in scene brightness at the pixel level \cite{gallego2020event, deng2021mvf, li2025semantic}.
Specifically, each pixel independently detects brightness variations in the logarithmic domain and triggers an event whenever the change exceeds a predefined threshold.
This sensing mechanism enables event cameras to achieve ultra-low latency (on the order of microseconds), low power consumption, and exceptionally high dynamic range (HDR), making them particularly well suited for fast motion capture and extreme lighting conditions \cite{rebecq2019events, ercan2023evreal, wu2024ev, wu2025dark, shao2024temporal}.

To better interpret event data, a fundamental task in event-based vision is event-to-video (E2V) reconstruction, which aims to recover intensity videos from raw event streams.
A wide range of learning-based approaches have been proposed, typically adopting recurrent encoder--decoder architectures that aggregate events over time and decode latent representations into intensity frames \cite{rebecq2019events, scheerlinck2020fast, stoffregen2020reducing, ercan2024hypere2vid, rebecq2019high}.
Despite steady progress, most prior works largely frame E2V as a temporal--spatial signal recovery problem, implicitly assuming that object-level structure and scene context can be recovered from event streams through sufficiently strong temporal aggregation and feature learning (Fig. \ref{fig:teaser}-b).
However, due to the change-driven sensing mechanism, event signals are inherently semantically under-determined: they lack rich semantic cues (Fig. \ref{fig:teaser}-a), such as object identity, texture, and global contextual information, that are naturally available in intensity frames \cite{tan2025spectrum, gallego2020event}.
As a result, existing approaches that rely solely on temporal and spatial modeling learn event feature representations that are inherently poor in semantic content and attention (Fig.~\ref{fig:teaser}-d-left).
Consequently, the reconstructed videos often remain semantically ambiguous at the object level, exhibiting distorted or inconsistent semantic details—e.g., inaccurate object boundaries or structural inconsistencies—especially in regions with weak or missing event evidence (Fig.~\ref{fig:teaser}-d-right).
In contrast, semantic information plays a critical role in video reconstruction and scene understanding, as it provides high-level structural constraints and contextual priors that help disambiguate the ill-posed mapping from sparse observations to dense intensity signals \cite{wang2024contextseg, liu2019structured, shocher2020semantic}.

From this perspective, we revisit the E2V task.
Rather than viewing E2V solely as a process of temporally aggregating sparse events and extracting spatial features via a simple encoder-decoder architecture, we argue that effective reconstruction requires explicitly modeling the missing semantic information inherent in event signals.
Accordingly, we propose a semantic-enriched end-to-end E2V training and supervision framework, called \textit{Semantic-E2VID}, with three key focuses: semantic abstraction, semantic feature fusion, and semantic-aware supervision (Fig. \ref{fig:teaser}-c).
Our approach enables the model to learn high-level semantic cues that are not explicitly captured by event sensors, integrate these cues into semantically enriched event representations (Fig. \ref{fig:teaser}-e-left), and leverage semantic-aware supervision to guide reconstruction toward structurally and semantically consistent video frames, even under sparse or ambiguous event conditions (Fig. \ref{fig:teaser}-e-right).
Specifically, to enable semantic abstraction for event data, we extract semantic knowledge from a pretrained vision foundation model, namely the Segment Anything Model (SAM) \cite{kirillov2023segment}, through a Cross-modal Feature Alignment (CFA) module.
This design avoids directly distilling frame-level representations into the event encoder, which can otherwise introduce modality-induced feature drift, and instead transfers semantic knowledge in a manner compatible with the event representation space.
To effectively exploit the learned semantics during reconstruction, we further integrate semantic cues into event feature representations via a Semantic-aware Feature Fusion (SFF) block.
This fusion strategy enables the decoder to jointly leverage temporal event information and semantic structure, allowing for a more faithful recovery of object boundaries and scene layout.
In addition, we introduce a Semantic Perceptual E2V Supervision that explicitly emphasizes semantic consistency during training by guiding the reconstruction process toward object-centric regions.
This supervision complements existing pixel-level and perceptual objectives and further reinforces the utilization of semantic cues during decoding.

Extensive experiments are conducted on a total of six public benchmarks, including three datasets with referenced ground-truth frames, two datasets with non-referenced frames, and one colorful reconstruction dataset.
Semantic-E2VID consistently achieves superior performance over previous state-of-the-art E2V approaches across multiple benchmarks and evaluation metrics, demonstrating clear advantages in both quantitative accuracy and qualitative reconstruction quality.
The key technical contributions of this work are summarized as follows:


\begin{itemize}
\item We revisit event-to-video reconstruction from a semantic perspective and propose Semantic-E2VID, a semantic-enriched E2V framework that explicitly addresses the semantic under-determinacy of event signals, going beyond conventional temporal and spatial modeling.
\item We introduce principled semantic modeling mechanisms for E2V that enable semantic abstraction, semantic feature fusion, and semantic-aware supervision, allowing event representation to capture object-level structure and contextual information lost in the sensing process.
\item Extensive experiments on six public benchmarks demonstrate that explicitly modeling and leveraging semantic information leads to more faithful reconstruction of object structure, clearer semantic boundaries, and improved robustness under challenging conditions.
\end{itemize}

\section{Related Work}
\noindent\textbf{Event-To-Video Reconstruction:}  
Event cameras capture scene dynamics by recording asynchronous brightness changes, offering advantages such as low latency, high dynamic range, and energy efficiency. One fundamental task in event-based vision is event-to-video (E2V) reconstruction, where the goal is to synthesize intensity frames from raw event streams. works \cite{rebecq2019events, scheerlinck2020fast, cadena2021spade, stoffregen2020reducing, ercan2024hypere2vid, weng2021event, ge2025eventmamba} have been done using either a recurrent-based, transformer-based, or Mamba-based network to learn event features over time for frame synthesis. More recent works \cite{liang2024e2vidiff, zhu2024temporal} leverage diffusion-based methods to tackle this problem; however, the time-consuming nature of diffusion models results in a heavy and inefficient forward process. Some \cite{liu2024seeing, zou2021learning} also explored HDR frame reconstruction. Despite these advancements, most existing E2V methods overlook semantic information such as object categories, texture, relationships, positions, etc, which are missing in the event modality yet important for video reconstruction. \cite{chen2024lase} leverages text description for semantic-aware E2V, but introduces text, as an extra input modality. Our proposed Semantic-E2VID is the first work that addresses this gap by distilling visual semantic knowledge from SAM to improve semantic learning in event modality and reconstruction quality.

\noindent\textbf{Learning Semantics From SAM:}
The Segment Anything Model (SAM) \cite{kirillov2023segment} has emerged as a powerful vision foundation model, demonstrating strong generalization capabilities across diverse vision tasks \cite{wang2024empirical, ge2025aquaslot}. SAM's success is attributed to its extensive pretraining on large-scale datasets, enabling it to produce robust and transferable semantic representations \cite{wang2025learn, gao2025cod}. Due to its strong semantic understanding, researchers have explored SAM for different downstream tasks. Works \cite{wu2025medical, ma2024segment, DENG2026112595, li2025domain} have first been done to transfer knowledge from SAM to medical image segmentation. Beyond that, some have utilized SAM as a teacher model for semantic knowledge distillation on other computer vision tasks, such as 2D Image Restoration \cite{zhang2024distilling} and feature learning and matching \cite{wu2025segment}. \cite{chen2023segment} distill knowledge from SAM for event-based segmentation. However, since both the teacher and student operate on the same task (segmentation), the distillation can be well aligned. In contrast, for the E2V task, directly distilling from frame-based models is challenging in transferring semantic knowledge from the frame modality to the event modality, as the objectives and data representations differ significantly \cite{messikommer2022bridging}. 


\section{Methodology}
\label{sec: method}
\subsection{Overview}
This section presents our semantic-aware event-to-video reconstruction framework. We begin by formulating the event-to-video task and introducing the input representation in Sec. \ref{sec:formulation}. We then discuss the established non-semantic event-to-video reconstruction baseline that serves as the spatiotemporal backbone in Sec. \ref{sec:baseline}. Building upon this backbone, we introduce Semantic-E2VID in Sec. \ref{sec:semantic-E2VID}, which explicitly injects high-level semantic abstraction into event representations through semantic learning and feature fusion. Finally, we present the semantic-aware supervision strategy in our end-to-end framework in Sec. \ref{sec:loss}.

\subsection{Modality and Task Formulation}
\label{sec:formulation}
Event sensors capture the algorithmic changes in the intensity. The output of an event sensor can be expressed as:

\begin{equation}
    E = \Gamma \left\{ \log \left( \frac{I_{t} + c}{I_{t-w} + c} \right), \epsilon \right\},
\label{Eq: event_formulation}
\end{equation}

where $E$ represents the recorded events, while $I_t$ and $I_{t-w}$ denote the intensity images at two different timestamps with a time window $w$. The function $\Gamma \left\{ \theta, \epsilon \right\}$ maps log-intensity changes to discrete event outputs, with $c$ serving as an offset to avoid $\log(0)$. The parameter $\epsilon$ defines the event triggering threshold. Specifically, $\Gamma\{\theta, \epsilon\} = 1$ when $\theta \geq \epsilon$, meaning a positive event triggered, and $\Gamma\{\theta, \epsilon\} = -1$ when $\theta \leq -\epsilon$, meaning a negative event triggered \cite{duan2021eventzoom, wu2025sweepevgs, liu2025event}.

In the event-to-video reconstruction pipeline, our objective is to reconstruct a sequence of images \(\{I_k\}\) from a continuous stream of events by proposing a model $M$, where each image \(I_k \in [0,1]^{W \times H}\) represents an intensity frame. Thus, the E2V task can be formulated into a simple process of $\{I_K\} = M(E)$. To accomplish this, we follow previous works \cite{rebecq2019events, rebecq2019high, ercan2024hypere2vid} to aggregate events into voxel grids for the adaptability of deep CNN architectures for processing event-based data. Let \( E_{T_k} \) represent a group of events spanning a time interval of \( \Delta T \) seconds, with \( T_k \) denoting the starting timestamp of this duration. The temporal resolution of the voxel grid is defined by \( B \), the number of discrete temporal bins used to quantize the timestamps of continuous-time events within the group. The resulting voxel grid \( V_k \in \mathbb{R}^{W \times H \times B} \) is constructed by first normalizing the event timestamps to the range \([0, B - 1]\), after which each event contributes its corresponding value to the appropriate voxel bin. $V_k$ will be the input of the first CNN encoder layers, and we use $B=5$ for all experiments.


\subsection{Spatiotemporal E2V Backbone}
\label{sec:baseline}
We first establish a non-semantic event-to-video reconstruction baseline that serves as a temporal-spatial backbone by following prior works~\cite{rebecq2019events, stoffregen2020reducing}.
Specifically, we adopt a U-Net-based recurrent encoder--decoder architecture originally proposed in~\cite{rebecq2019events}, which serves as a representative backbone for modeling temporal dynamics and spatial structures in event streams.
The encoder consists of a sequence of convolutional layers and ConvLSTM units, where each convolutional layer uses a kernel size of $5$ with a stride $2$, and each ConvLSTM employs a kernel size of $3$ to preserve spatial and channel consistency between inputs and hidden states.
Residual blocks are constructed using two convolutional layers with a kernel size $3$.
Given an input event voxel grid $V_k \in \mathbb{R}^{W \times H \times B}$, the encoder extracts event features $\mathcal{F}_e \in \mathbb{R}^{\frac{1}{2}W \times \frac{1}{2}H \times 256}$. The decoder mirrors the encoder structure and is composed of bilinear upsampling followed by convolutional layers, progressively reconstructing intensity frames from the encoded event features.
Additionally, following~\cite{ercan2024hypere2vid}, the decoder incorporates auxiliary context fusion mechanisms, which dynamically generate convolutional weights conditioned on both current event inputs and previously reconstructed frames to enhance temporal coherence.

Despite their effectiveness, such architectures fundamentally focus on temporal and spatial modeling of event data and do not explicitly address the semantic under-determinacy inherent in event signals.
As a result, the learned event representations remain limited in semantic content, making it difficult to recover object-level structure and contextual information from sparse event observations.

\begin{figure*}[h]
	
	\centering
	
	\includegraphics[width=\linewidth,scale=1.0]{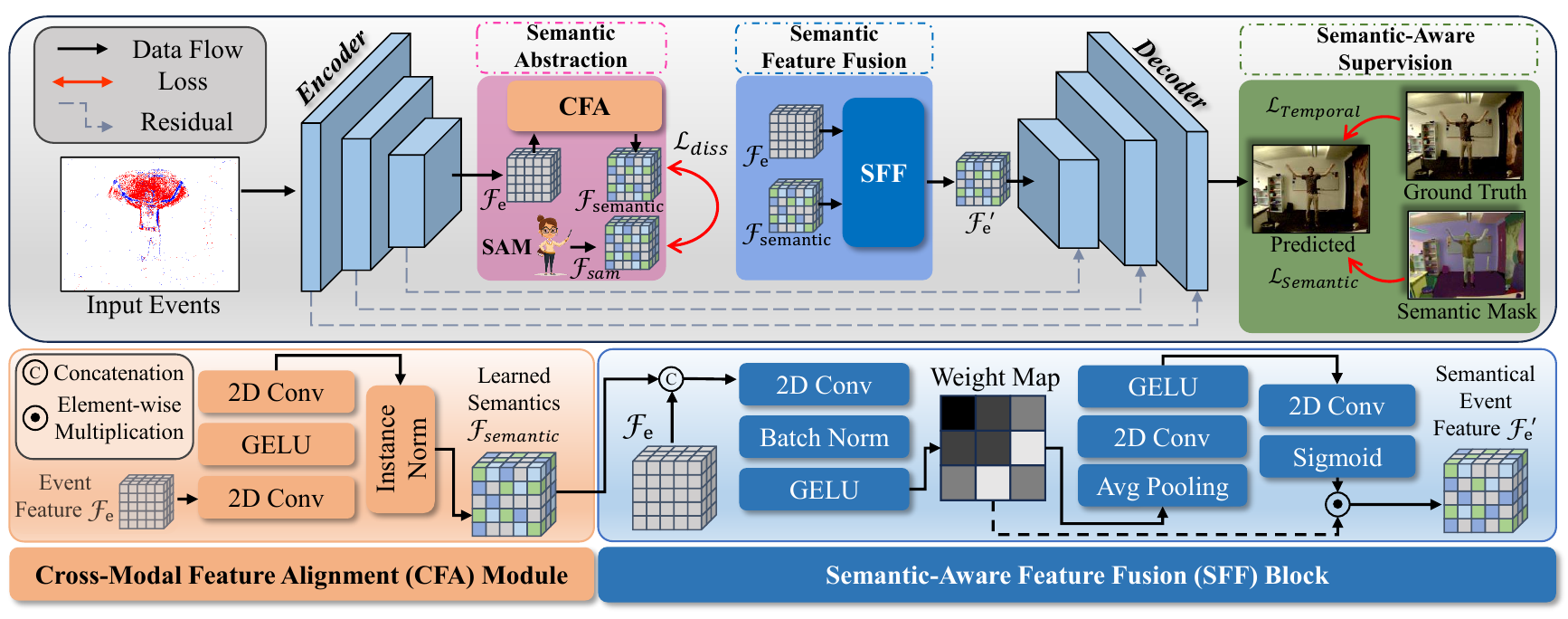}
	
	\caption{
Overview of the proposed \textit{Semantic-E2VID} framework.
Given input events, an encoder first extracts event features $\mathcal{F}_e$.
To enable \emph{semantic abstraction} without introducing modality-induced feature drift, we introduce the Cross-Modal Feature Alignment (CFA) module that aligns $\mathcal{F}_e$ with frame semantic features $\mathcal{F}_{sam}$ extracted from a pretrained vision foundation model (SAM), yielding learned frame semantic representations $\mathcal{F}_{semantic}$.
To effectively inject semantic knowledge into the event representation space, the Semantic-Aware Feature Fusion (SFF) block integrates $\mathcal{F}_{semantic}$ with $\mathcal{F}_e$, producing semantically enriched event features $\mathcal{F}_e'$.
A decoder then reconstructs the final intensity frame from $\mathcal{F}_e'$.
During training, the reconstruction process is jointly supervised by conventional reconstruction and temporal losses, together with semantic-aware supervisions that emphasize semantic consistency in object-centric regions.
}

	\label{fig:pipeline}
	
\end{figure*}

\subsection{Semantic-E2VID}
\label{sec:semantic-E2VID}
Motivated by this limitation, we revisit the E2V task from a semantic perspective. Rather than treating reconstruction purely as a temporal-spatial signal recovery problem, we reformulate E2V as a \emph{semantic-enriched reconstruction task}, where effective reconstruction requires explicitly learning, fusing, and supervising high-level semantic information that is absent in event data.
In particular, effective semantic-enriched E2V necessitates three key capabilities:
(i) \emph{semantic abstraction}, i.e., extracting high-level semantic knowledge from complementary modalities and aligning it with event representations; (ii) \emph{semantic feature fusion}, i.e., integrating learned semantics with encoded event features in a manner compatible with the event representation space; and (iii) \emph{semantic-aware supervision}, i.e., guiding the reconstruction process toward object-level semantic consistency.

To realize these capabilities in an end-to-end manner, we propose \textit{Semantic-E2VID}, a semantic-guided E2V framework composed of three corresponding components: Cross-Modal Feature Alignment (CFA) module for semantic abstraction, Semantic-aware Feature Fusion (SFF) block for injecting semantics into event features, and Semantic Perceptual E2V Supervision strategy for enforcing semantic consistency during reconstruction. An overview of the framework is shown in Fig.~\ref{fig:pipeline}, and each component is detailed in the following subsections.

\subsubsection{Semantic Abstraction and Learning.}
The goal of CFA is to explore semantic information by learning valuable knowledge with rich semantics from SAM. However, a natural modality mismatches as opposed to direct distillation. That is: the event feature $\mathcal{F}_e \in \mathbb{R}^{\frac{1}{2} W \times \frac{1}{2} H \times 256}$ encoded by the event encoder represent completely different information compared to the encoded frame feature with rich semantics $\mathcal{F}_{sam} \in \mathbb{R}^{\frac{1}{2} W \times \frac{1}{2} H \times 256}$ by SAM\footnote{Note: SAM's feature is extracted from the ground truth grayscale frames of the training dataset by SAM's image encoder and was interpolated to the corresponding spatial dimension. No extra computational cost is needed during the inference stage. This also holds for the feature distillation learning process.}. Therefore, directly applying the distillation loss between $\mathcal{F}_e$ and $\mathcal{F}_{\text{sam}}$ will cause a shifting in the event encoder learning and decoding process, which results in poor reconstruction, as shown by the ablation study in later sections. Thus, CFA is proposed to bridge the misalignment gap by mapping the encoded event feature $\mathcal{F}_e$ to SAM's semantics without interfering with the event encoder. Specifically, as shown in Fig. \ref{fig:pipeline} orange box, CFA takes the encoded event feature $\mathcal{F}_e$ as input and applies two 2D convolution operations on event features, combined with the GELU activation function, to map local event features to local frame features from SAM with rich semantics, and enhance semantic representation capabilities. Subsequently, Instance Norm is utilized to normalize features and align them with the statistical distribution of semantic features, reducing the impact of distribution discrepancies. Finally, to ensure the model can effectively utilize semantic information while preventing the propagation of error information, we employ distillation loss, which will be explained in a later section, to constrain the learned semantic features $\mathcal{F}_{\text{semantic}} \in \mathbb{R}^{\frac{1}{2} W \times \frac{1}{2} H \times 256}$ to approximate the ideal semantic features $\mathcal{F}_{\text{sam}}$ provided by SAM while avoiding misleading event encoder learning.

\subsubsection{Semantic Feature Fusion.}
While CFA enables Semantic-E2VID to explore rich semantic information, the next challenge is to fuse the learned semantics $\mathcal{F}_{\text{semantic}}$ with the encoded event feature $\mathcal{F}_e$ to generate encoded event features with rich semantics, ready for decoding into an accurate video. To conquer this, SFF is employed, where its primary function is to enable deep interaction between $\mathcal{F}_{\text{semantic}}$ and $\mathcal{F}_e$ through a spatial attention mechanism and an adaptive fusion strategy, ensuring that semantic information exerts its influence at appropriate spatial locations. Specifically, as shown in Fig. \ref{fig:pipeline} blue box, SFF first concatenates $\mathcal{F}_{\text{semantic}}$ from CFA with the original event features $\mathcal{F}_e$. Subsequently, 2D convolution, Batch Normalization, and GELU activation functions are employed to transform the features into an attention-like weight map. To ensure that semantic information acts at the correct position, we employ a self-attention mechanism \cite{vaswani2017attention} where global context information of the entire feature map is extracted via global average pooling, followed by actions such as 2D convolution and sigmoid activation functions, respectively. Finally, based on the attention weights, element-wise weighting is performed, and the adaptively generated semantic event features $\mathcal{F}_{\text{e}}^{\prime}$ is fed to later decoders for image reconstruction.

\begin{figure}[h]
\vspace{-5pt}
\centering
\includegraphics[width=0.6\linewidth]{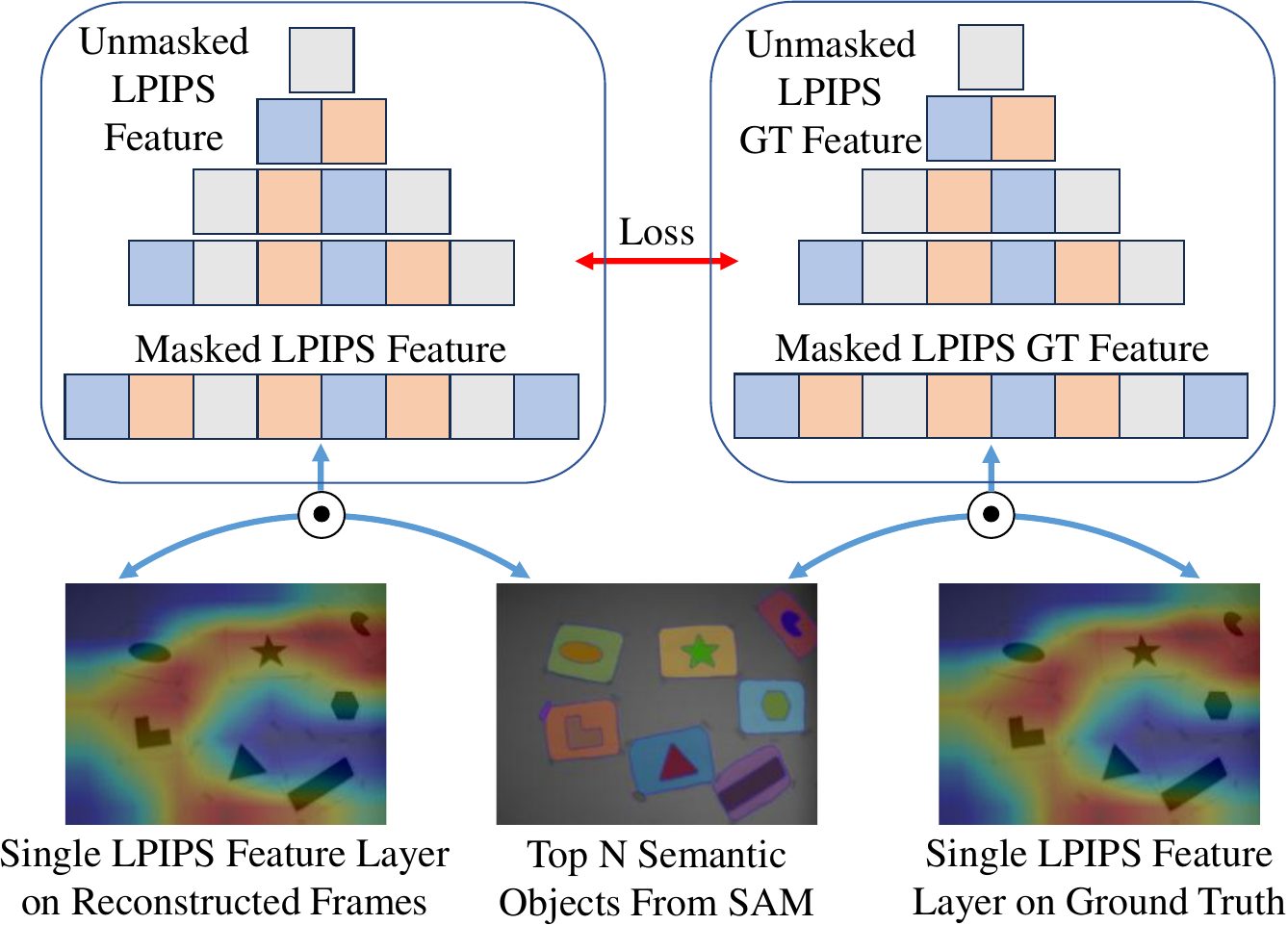}
\vspace{-5pt}
\caption{Illustration of our proposed Semantic Perceptual E2V Supervision. Instead of computing the perceptual loss uniformly over the entire image, we apply a binary categorical mask $\mathcal{M}$ derived from SAM’s segmentation outputs to the first-layer features of the perceptual network (AlexNet). This enhances supervision on semantically important regions (e.g., objects), while deeper layers retain full-frame guidance to preserve overall image quality.}
\label{fig:enhance_lpips}
\vspace{-5pt}
\end{figure}

\subsection{Semantic-Aware Supervision Event To Video Supervision}
\label{sec:loss}
Existing supervisions used in current E2V works often only consider the pixel-level reconstruction \cite{ma2024color4e} or perceptual-level reconstruction but ignore the importance of regional semantic information \cite{rebecq2019high, ercan2024hypere2vid}. This may lead to an inaccurate reconstruction of key areas with rich semantics (such as shapes, edges, and relationships of many key objects). Therefore, we proposed a novel Semantic Perceptual E2V Supervision, along with the widely adopted Temporal Consistency Loss by existing E2V works and Relational Knowledge Distillation Loss for learning semantic information from SAM, to ensure accurate reconstruction. 

\subsubsection{Semantic Perceptual E2V Supervision.} The traditional perceptual reconstruction loss passes the reconstructed image and reference images through a pre-trained deep neural network trained for visual recognition tasks and uses the distance between the multiple layers of deep features from the network as a measure of the perceptual difference between the two images. However, the perception was done on the entire reconstructed frames, which may lead the E2V model to focus excessively on larger background regions and ignore small but critical objects with key semantics. To address this challenge, we utilize segmentation category results from SAM by employing its mask generation function\footnote{https://github.com/facebookresearch/segment-anything} to produce fine-grained semantic groupings. Specifically, single-point prompts are sampled in a grid across the image, from which SAM predicts multiple candidate masks. These masks are subsequently filtered for quality and deduplicated using non-maximum suppression~\cite{kirillov2023segment}. Finally, we select the top $N$ non-background categories and resize them to construct a binary categorical mask $\mathcal{M} \in \mathbb{R}^{h \times w \times N}$. As illustrated in Fig.~\ref{fig:enhance_lpips}, to better supervise the key regions with rich semantics, we dot multiply the first, largest layer feature of the LPIPS features with $\mathcal{M}$ ($h=w=27$ to match the spatial size), to derive a masked LPIPS semantic feature layer. The remaining layers of the LPIPS feature remain unmasked to maintain the reconstruction quality of the background regions.
Following previous E2V works \cite{ercan2024hypere2vid, rebecq2019high}, we use the AlexNet \cite{krizhevsky2017imagenet} as the network for perceptual image patch similarity (LPIPS) \cite{zhang2018unreasonable}, and we use the value $10$ for $N$ for all experiments. The Semantic Perception Supervision $\mathcal{L}_{\text{Semantic}}$ is calculated as below:

\begin{equation}
\begin{aligned}
\mathcal{L}_{\text{Semantic}} = &\left\| \mathcal{M} \odot \left( \phi_1(I_{\text{rec}})_0 - \mathcal{M} \odot \phi_1(I_{\text{gt}} \right)_0 \right\|_2 \\ + 
&\text{LPIPS}(\phi_l(I_{\text{rec}})_{1:} - \phi_l(I_{\text{gt}})_{1:})
\end{aligned}
\end{equation}

where $\odot$ represents the element-wise multiplication operation, $I_{\text{rec}}$ and $I_{\text{gt}}$ are reconstructed frame and ground truth frame, $\phi_l(\cdot)$ represents AlexNet, the footnote of $_0$ represent the first, largest layer, the footnote of $_{1:}$ represent the remaining layers, and the $\left\| \cdot \right\|_2$ denotes the $L_2$ distance.

\subsubsection{Learning Semantics via Distillation.}
We adopt relational knowledge distillation similar to previous distillation works \cite{wu2025segment, park2019relational} where the learned semantics $\mathcal{F}_{\text{semantic}}$ and semantics from SAM $\mathcal{F}_{\text{sam}}$ are flattened by their spatial dimension into feature maps \( \mathcal{V}_{semantic} \in \mathbb{R}^{H \times W, C} \) and \( \mathcal{V}_{sam} \in \mathbb{R}^{H \times W, C} \), after normalizing each feature map along the channel dimension. The spatial similarity matrix of students or teachers is calculated by inner product and the Distillation Loss can be calculated by the $L_1$ loss between the similarity matrices, as shown below:
\begin{equation}
\mathcal{L_{\text{diss}}} = \left\|  {\mathcal{V}_{semantic}^{T} \cdot \mathcal{V}_{semantic}} - {\mathcal{V}_{sam}^{T} \cdot \mathcal{V}_{sam}} \right\|_1,
\end{equation}


\subsubsection{Ensuring Temporal Consistency.}
Following previous E2V works \cite{rebecq2019high, ercan2024hypere2vid}, we deploy temporal consistency loss between the reconstructed images in adjacent time steps of the network. Specifically, temporal consistency loss is achieved by warping the previous reconstructed image using a ground truth optical flow to align with the current reconstructed image, and using the masked distance between these aligned images as a weight for temporal consistency, where the warping error between the previous true intensity image and the current true intensity image calculates the masked distance. The temporal consistency loss is calculated as:
\begin{equation}
\mathcal{L}_k^{\text{TC}} = M_k \left\| \hat{I}_k - W\left(\hat{I}_{k-1}, F_{k \to k-1} \right) \right\|_1,
\end{equation}
where $F_{k \to k-1}$ denotes the optical flow map between time
steps k and k-1, W is the warping function, and $M_k$ represents the occlusion mask which is computed as:
\begin{equation}
M_k = \exp\left(-\alpha \left\| I_k - W\left(I_{k-1}, F_{k \to k-1} \right) \right\|_2^2 \right).
\end{equation}


\subsubsection{Final Optimization.} Combining the three losses, our final optimization objective is:
\begin{equation}
\mathcal{L} = \mathcal{L}_{\text{Semantic}} + \mathcal{L}_{\text{temporal}} + \lambda \mathcal{L}_{\text{diss}}.
\label{eq: final_loss}
\end{equation}

\section{Experiments}
This section presents the experimental setup and evaluation of our proposed method. We first describe the \textit{Training Protocols} and \textit{Evaluation Protocols}, detailing implementation settings and benchmark configurations. We then provide a comprehensive \textit{Qualitative Analysis} to visually assess reconstruction quality across various scenarios, followed by an in-depth \textit{Ablation Study} that quantifies the contribution of each proposed component and analyzes model robustness.

\label{sec: exps}
\subsection{Training Protocols}
\subsubsection{Training Dataset.} We randomly sampled images from the MSCOCO dataset \cite{lin2014microsoft} and utilized multi-object 2D renderer option of ESIM \cite{rebecq2018esim} to generate synthetic training datasets following \cite{stoffregen2020reducing}. These images serve as the background and foreground objects for the dataset. The dataset comprises $280$ sequences, each lasting $10$ seconds. Within the sequences, foreground images exhibit varying speeds and trajectories. Each sequence includes the generated event stream, ground truth images, and optical flow maps with an average rate of 51 Hz. Both the event camera and the frame camera have a resolution of $256 \times 256$ pixels, while the contrast threshold range for event generation is $0.1$ to $1.5$.

\subsubsection{Implementation Details.}
We utilized sequences of length $40$ to train our network, initializing the network parameters using He initialization. We set the batch size to 4 and trained the model for 400 epochs using the Adam optimizer, with a learning rate of $1 \times e^{-5}$. We use $\lambda=1.8$, and $\alpha = 50$ following \cite{ercan2024hypere2vid, rebecq2019high, lai2018learning} as hyper-parameters for the loss function.
All networks were implemented in PyTorch and executed on a system equipped with a single RTX $3090$ GPU.

\subsection{Evaluation Protocols}
To comprehensively evaluate the effectiveness of our approach, we employ three types of evaluation protocols: (\textit{i}) quantitative comparison for data with referenced ground truth frames; (\textit{ii}) quantitative comparison for data with non-referenced frames; and (\textit{iii}) qualitative comparison with color frame reconstruction. We use six real-world datasets in total. Event Camera Dataset \cite{mueggler2017event} (ECD), the Multi-Vehicle Stereo Event Camera dataset \cite{zhu2018multivehicle} (MVSEC), and the High-Quality Frames dataset (HQF) \cite{stoffregen2020reducing}, are used for evaluation with referenced GT frames, deploying mean squared error (MSE), structural similarity
\cite{wang2004image} (SSIM), and learned perceptual image patch similarity \cite{zhang2018unreasonable} (LPIPS) as evaluation metrics. 
The two datasets, high-speed and HDR sequences from \cite{rebecq2019high}, and the NIGHT dataset in MVSEC, are used for evaluation on reconstruction with no referenced GT frames. Following previous works \cite{ercan2024hypere2vid, ercan2023evreal}, we use the NIQE \cite{mittal2012making} metric for non-reference evaluation. For colorful frames reconstruction, we use CED and compare the results qualitatively.

We perform a comprehensive comparison against state-of-the-art representative methods across all different evaluation protocols. To ensure fairness, we utilize the official pre-trained models released by the respective authors and evaluate them under their default experimental settings. 

\subsubsection{Evaluation on Data with Referenced Frames.}
Table \ref{tab:ecd_mvsec_horizontal} reports the quantitative results obtained using the aforementioned evaluation metrics. Our method consistently achieves state-of-the-art performance across most metrics. On the ECD and HQF datasets, it significantly outperforms HyperE2VID and EventMamba. For the MVSEC dataset, our approach attains the lowest MSE and highest SSIM scores, while also achieving competitive LPIPS results compared to transformer-based ET-Net with heavy computation cost \cite{ercan2024hypere2vid}. These findings highlight the effectiveness of Semantic-E2VID in reconstructing high-fidelity images. By incorporating semantic priors during both feature encoding and decoding, our model can more accurately reconstruct structural and contextual information that corresponds to meaningful visual semantics in the scene.

\begin{table*}[t]
\centering
\small
\caption{Quantitative comparison on ECD, MVSEC and HQF datasets. Lower is better ($\downarrow$) for MSE and LPIPS. Higher is better ($\uparrow$) for SSIM. The best and second-best indexes are marked in bold and \underline{underlined}.}
\resizebox{\textwidth}{!}{
\begin{tabular}{lccccccccccc}
\toprule
\textbf{Method} & \multicolumn{3}{c}{\textbf{ECD}} & \multicolumn{3}{c}{\textbf{MVSEC}} & \multicolumn{3}{c}{\textbf{HQF}} \\
\cmidrule(lr){2-4} \cmidrule(lr){5-7} \cmidrule(lr){8-10}
& MSE$\downarrow$ & SSIM$\uparrow$ & LPIPS$\downarrow$
& MSE$\downarrow$ & SSIM$\uparrow$ & LPIPS$\downarrow$
& MSE$\downarrow$ & SSIM$\uparrow$ & LPIPS$\downarrow$\\
\midrule
E2VID \cite{rebecq2019high}  & 0.179 &  0.450 & 0.322 & 0.225 &  0.241 & 0.645 &  0.099 &  0.462  &  0.388\\
FireNet \cite{scheerlinck2020fast} & 0.133 &  0.459 & 0.321 & 0.294 &  0.198 & 0.702  &  0.100 &  0.422  &  0.463\\
E2VID+ \cite{stoffregen2020reducing} & 0.070 &  0.503 & 0.236 & 0.132 &  0.262 & 0.514 &  0.036 &  0.533  &  \underline{0.252}\\
FireNet+ \cite{stoffregen2020reducing} & 0.062 &  0.452 & 0.289 & 0.219 &  0.212 & 0.570 &  0.040 &  0.471  &  0.314\\
SPADE-E2VID \cite{cadena2021spade} & 0.091 &  0.461 & 0.337 & 0.138 &  0.266 & 0.591 & 0.079 &  0.405  &  0.514\\
SSL-E2VID \cite{paredes2021back}  & 0.092 &  0.415 & 0.380 & 0.124 &  0.264 & 0.694 &  0.090 &  0.407  &  0.496\\
ET-Net \cite{weng2021event} & 0.047 &  0.552 & 0.224 & 0.107 &  0.288 & 0.489 &  0.034 &  0.534  &  0.268\\
HyperE2VID \cite{ercan2024hypere2vid} & \underline{0.033} &  \underline{0.576} & \underline{0.212} & 0.076 &  0.315 & \underline{0.477} &  0.032 &  0.531  &  0.260\\
STLR \cite{tang2024spike} & - & - & - & 0.128 & 0.251 & 0.552 & 0.089 & 0.349 & 0.408 \\
EventMamba \cite{ge2025eventmamba} & - & - & - & \underline{0.073} & \underline{0.328} & \textbf{0.475} & \underline{0.031} & \underline{0.575} & 0.261 \\
\midrule
baseline & 0.035 &  0.560 & 0.221 & 0.107 & 0.291 & 0.506  &  0.035 &  0.520  &  0.283 \\
\textbf{Ours} & \textbf{0.014} &  \textbf{0.594} & \textbf{0.208} & \textbf{0.067} &  \textbf{0.329} & 0.483 &  \textbf{0.030} &  \textbf{0.577}  &  \textbf{0.219} \\
\bottomrule
\end{tabular}
}
\label{tab:ecd_mvsec_horizontal}
\end{table*}

\subsubsection{Evaluation on Data with no Referenced GT Frames.}
Table \ref{tab:HDR_and_NIGHT} presents the quantitative results obtained from evaluating the methods on sequences from the HDR and NIGHT datasets. These benchmarks showcase the reconstruction ability under challenging conditions such as low light, overexposure, and extremely fast motions, where ground truth frames are not reliable. Our method achieves the best NIQE score in both benchmarks with challenging scenarios involving fast camera motion. This robustness stems from the learned semantic guidance, which helps maintain coherent object appearance and scene layout even when low-level signal quality is degraded.

\begin{table}[h]
\centering
\scriptsize
\setlength{\tabcolsep}{5mm} 
\caption{NIQE metric ($\downarrow$) on HDR and NIGHT datasets. Best and second-best are in bold and \underline{underline}.}
\begin{tabular}{lcc}
\toprule
\textbf{Method} & \textbf{HDR} & \textbf{NIGHT} \\
\midrule
E2VID \cite{rebecq2019high}     & 4.347  & 6.719  \\
FireNet \cite{scheerlinck2020fast}   & 4.087  & 6.319  \\
SPADE-E2VID \cite{cadena2021spade} & 5.570  & 7.525  \\
SSL-E2VID \cite{paredes2021back} & 6.168  & 12.037 \\
HyperE2VID \cite{ercan2024hypere2vid} & \underline{3.758} & \underline{6.295} \\        
\midrule
\textbf{Ours} & \textbf{3.647} & \textbf{5.968} \\
\bottomrule
\end{tabular}
\label{tab:HDR_and_NIGHT}
\end{table}

\subsubsection{Evaluation on Color Frame Reconstruction.} 
\begin{wrapfigure}{r}{0.6\linewidth}
\vspace{-10pt}
\centering
\includegraphics[width=1\linewidth]{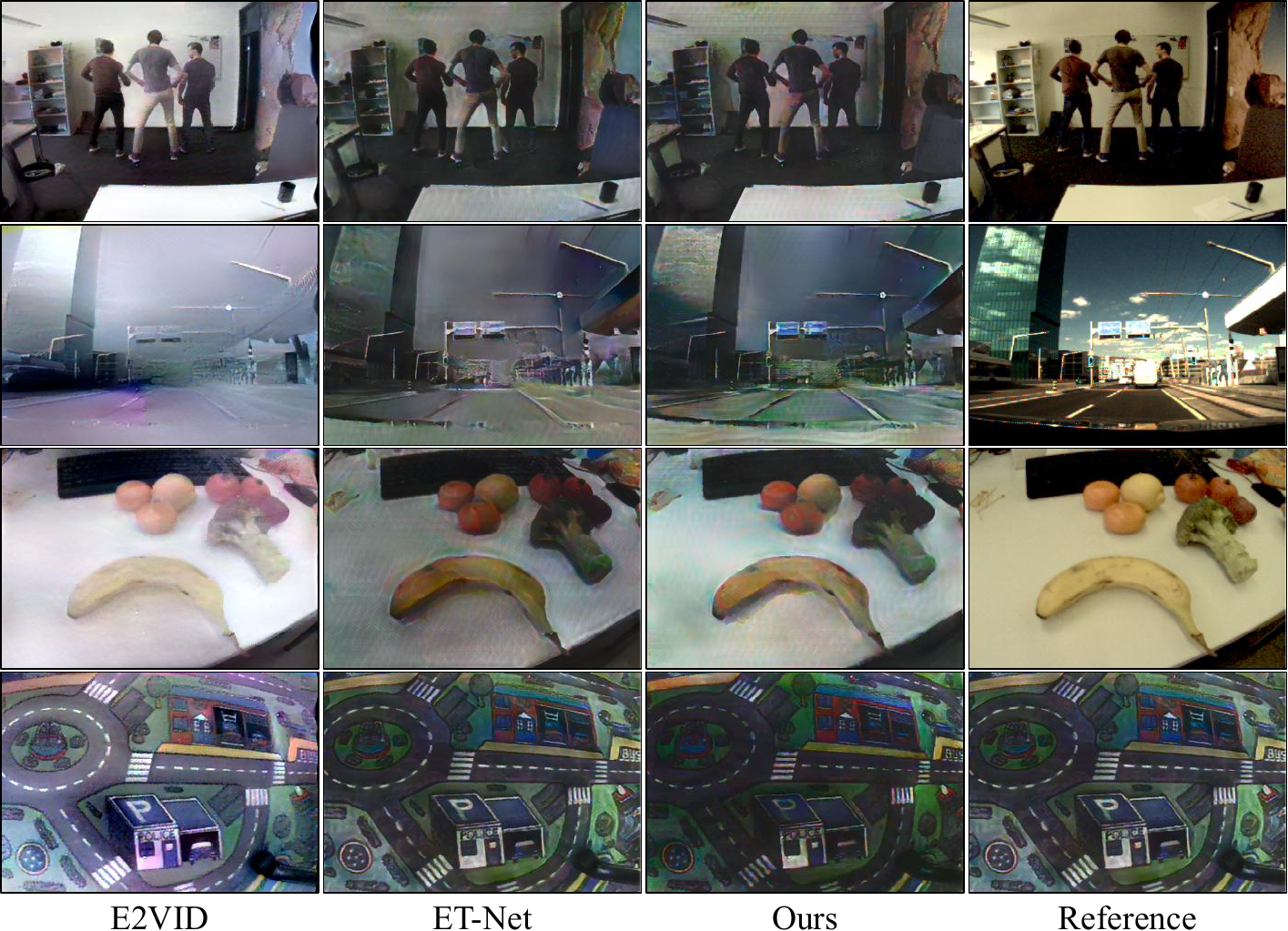}
\vspace{-10pt}
\caption{Visual comparison for colorful frame reconstruction on the CED dataset. 
    Our approach demonstrates better color fidelity and structural consistency, effectively recovering fine-grained details and natural color tones even under challenging conditions.}
\label{fig:color}
\vspace{-10pt}
\end{wrapfigure}

To further demonstrate the ability of our method for color image reconstruction, we use the CED dataset, which utilizes the DAVIS 346c camera to capture color frames and events. This dataset is essential for demonstrating the color reconstruction capability of the proposed method. As shown in Fig. \ref{fig:color}, Semantic-E2VID produces more visually compelling reconstructions, particularly in scenes containing colored objects and fast-moving dynamics. The reconstructed images of Semantic-E2VID have higher contrast and clear edges, which proves the effectiveness and robustness of our proposed method.

\subsubsection{Evaluation on Inference Speed.}
\begin{figure*}[h]
  \centering
  \includegraphics[width=\linewidth]{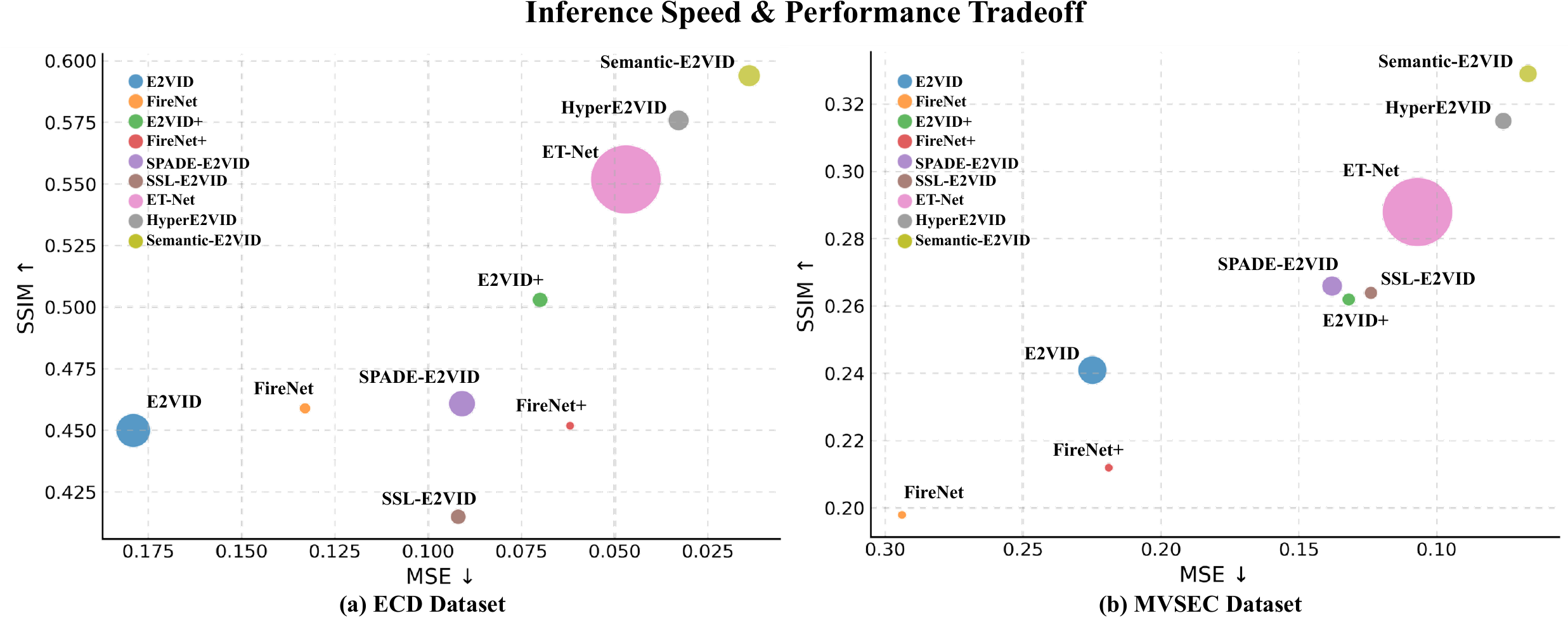} 
  \caption{Inference speed--performance tradeoff for event-to-video reconstruction on two benchmarks:
(a) ECD and (b) MVSEC.
Each marker corresponds to an E2V method, where the horizontal axis denotes MSE (lower is better, leftward)
and the vertical axis denotes SSIM (higher is better, upward).
The area of each circle is proportional to the inference speed (FPS), indicating computational efficiency,
while different colors represent different methods.
Methods located toward the upper-left achieve better reconstruction quality,
and larger circles indicate faster inference.
Overall, \textit{Semantic-E2VID} consistently occupies the most favorable region,
achieving lower reconstruction error, higher structural similarity, and a superior accuracy--efficiency tradeoff
compared to prior approaches.
  } 
  \label{fig:speed}
\end{figure*}

In addition to reconstruction accuracy, inference efficiency is a critical factor for E2V reconstruction, especially in real-time and high-speed vision applications.
To evaluate the tradeoff between reconstruction performance and inference speed, we compare Semantic-E2VID with representative E2V methods in terms of both accuracy and frames per second (FPS) on two datasets.
Fig.~\ref{fig:speed} visualizes this tradeoff by jointly plotting reconstruction quality and computational efficiency.

Specifically, the horizontal axis represents the reconstruction error measured by MSE (lower is better), while the vertical axis denotes structural similarity measured by SSIM (higher is better).
Inference speed FPS is encoded by the area of each marker, where larger circles indicate faster processing.
Methods located toward the upper-left region therefore achieve a more favorable balance between reconstruction accuracy and efficiency.

As shown in Fig.~\ref{fig:speed}, Semantic-E2VID consistently occupies the most desirable region across both ECD and MVSEC datasets.
Compared to prior approaches, our method achieves significantly lower reconstruction error and higher structural similarity, while maintaining competitive inference speed.
Despite incorporating additional semantic modeling components, Semantic-E2VID does not sacrifice efficiency and remains suitable for practical deployment.
These results demonstrate that explicitly modeling semantic information improves reconstruction quality without introducing prohibitive computational overhead, leading to a superior accuracy--efficiency tradeoff.

\subsection{Qualitative Analysis}
To further validate the effectiveness and generalization ability of our approach, we present comprehensive qualitative results across various settings. We examine the reconstruction quality of semantically meaningful regions, the effectiveness of our method in enabling downstream semantic segmentation, and the interpretability of learned semantic features. These results collectively demonstrate the semantic awareness and visual fidelity achieved by our framework.

\subsubsection{Overall Qualitative Comparison.}
\begin{figure*}[h]
  \centering
  \includegraphics[width=\linewidth]{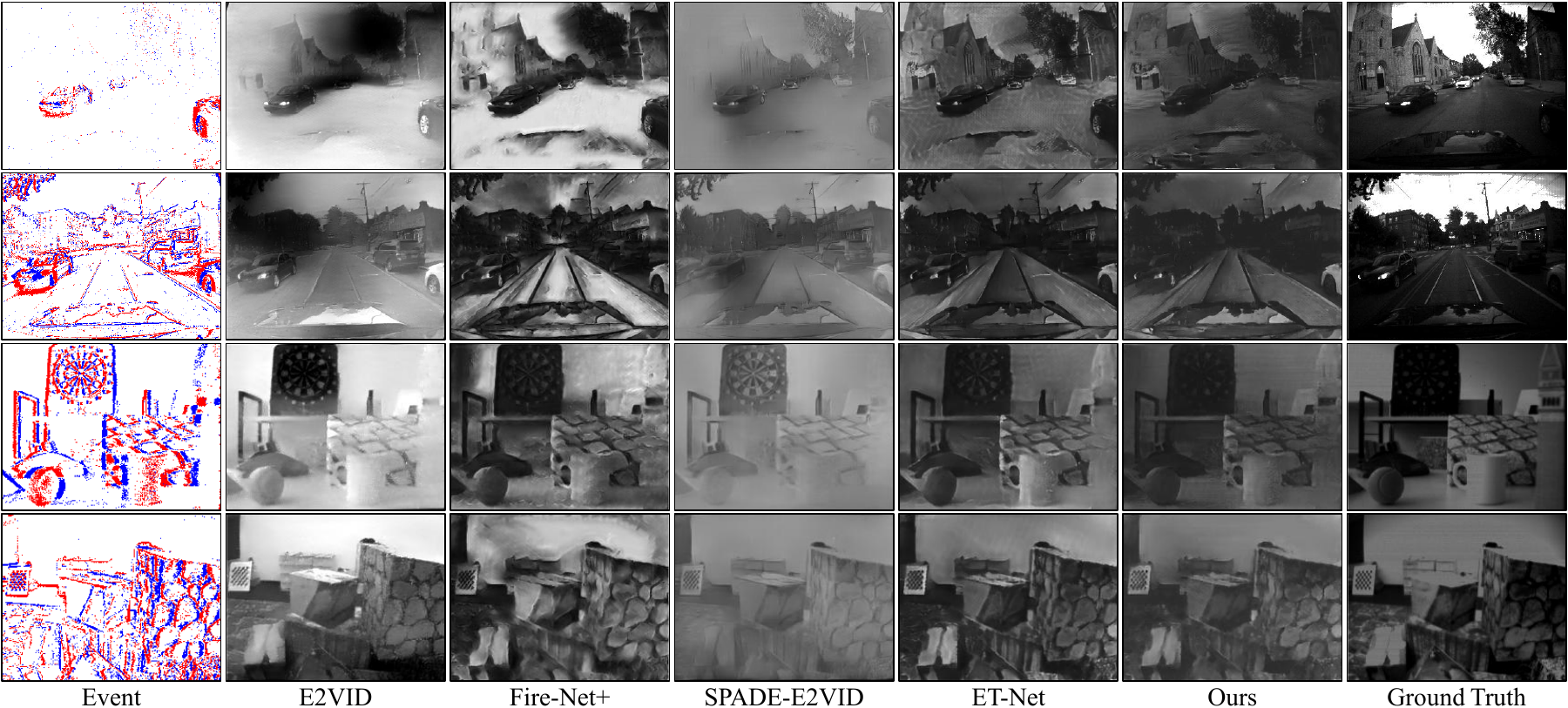} 
  \caption{Qualitative comparisons on sequence datasets with referenced GT frames. 
    Semantic-E2VID produces sharper textures, more accurate structural details, and fewer artifacts compared to other methods, particularly under challenging illumination and motion conditions. 
    The integration of high-level semantic priors enables our approach to better reconstruct complex scenes and recover missing visual cues from sparse event data.
  } 
  \label{fig:visualize}
\end{figure*}

Fig. \ref{fig:visualize} showcases qualitative reconstruction results of our Semantic-E2VID alongside all other SOTA methods on samples from datasets with referenced GT frames. It can be observed that methods such as E2VID, FireNet+, and SPADE-E2VID struggle with brightness consistency, leading to noticeably degraded visual quality. Our Semantic-E2VID demonstrates superior performance in preserving fine-grained object boundaries. This improvement is attributed to the incorporation of rich semantic information, which enhances the overall fidelity of the reconstruction. Furthermore, the outputs of Semantic-E2VID show a closer resemblance to the ground truth frames, delivering high visual quality while effectively mitigating reconstruction artifacts. Semantic-aware guidance allows our model to reconstruct high-level scene understanding, including object shapes and spatial relationships, yielding more realistic and interpretable video frames.

\begin{figure*}[h!]
  \centering
  \includegraphics[width=\linewidth]{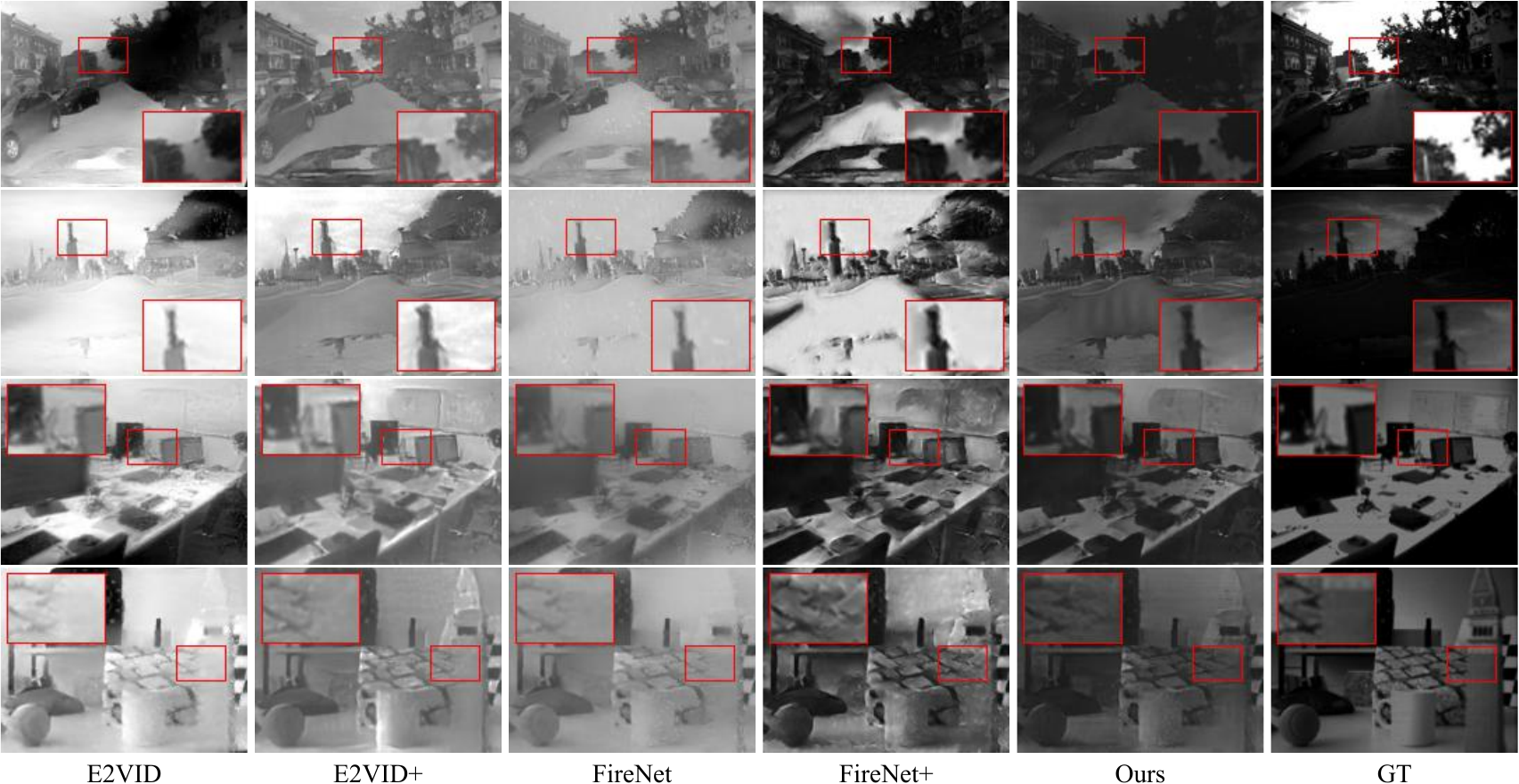} 
  \caption{Visual illustration of reconstruction quality across multiple datasets. Red boxes highlight semantically important regions where our method outperforms others in preserving structure and detail.
  } 
  \label{fig:visualize_red_box}
\end{figure*}

\subsubsection{Qualitative Analysis on Reconstruction Quality of Semantic Regions}
To further demonstrate the effectiveness of our method in reconstructing semantically meaningful regions, we provide qualitative comparisons with state-of-the-art methods, including E2VID~\cite{rebecq2019high}, E2VID+~\cite{stoffregen2020reducing}, FireNet~\cite{scheerlinck2020fast}, FireNet+~\cite{stoffregen2020reducing}, and ours. As shown in Fig.~\ref{fig:visualize_red_box}, our method produces more accurate and sharper reconstructions in regions with rich semantic content (e.g., pedestrians, vehicles, or structured indoor objects).

Compared to other approaches, which often suffer from blurred contours, missing parts, or distorted semantic structures, our method preserves fine-grained semantic information and spatial details more effectively. This improvement highlights the benefit of our semantic-aware reconstruction strategy, which guides the network to focus more on regions that are semantically important and perceptually significant.

\subsubsection{Qualitative Analysis on Downstream Segmentation Task}
\begin{figure*}[h]
  \centering
  \includegraphics[width=\linewidth]{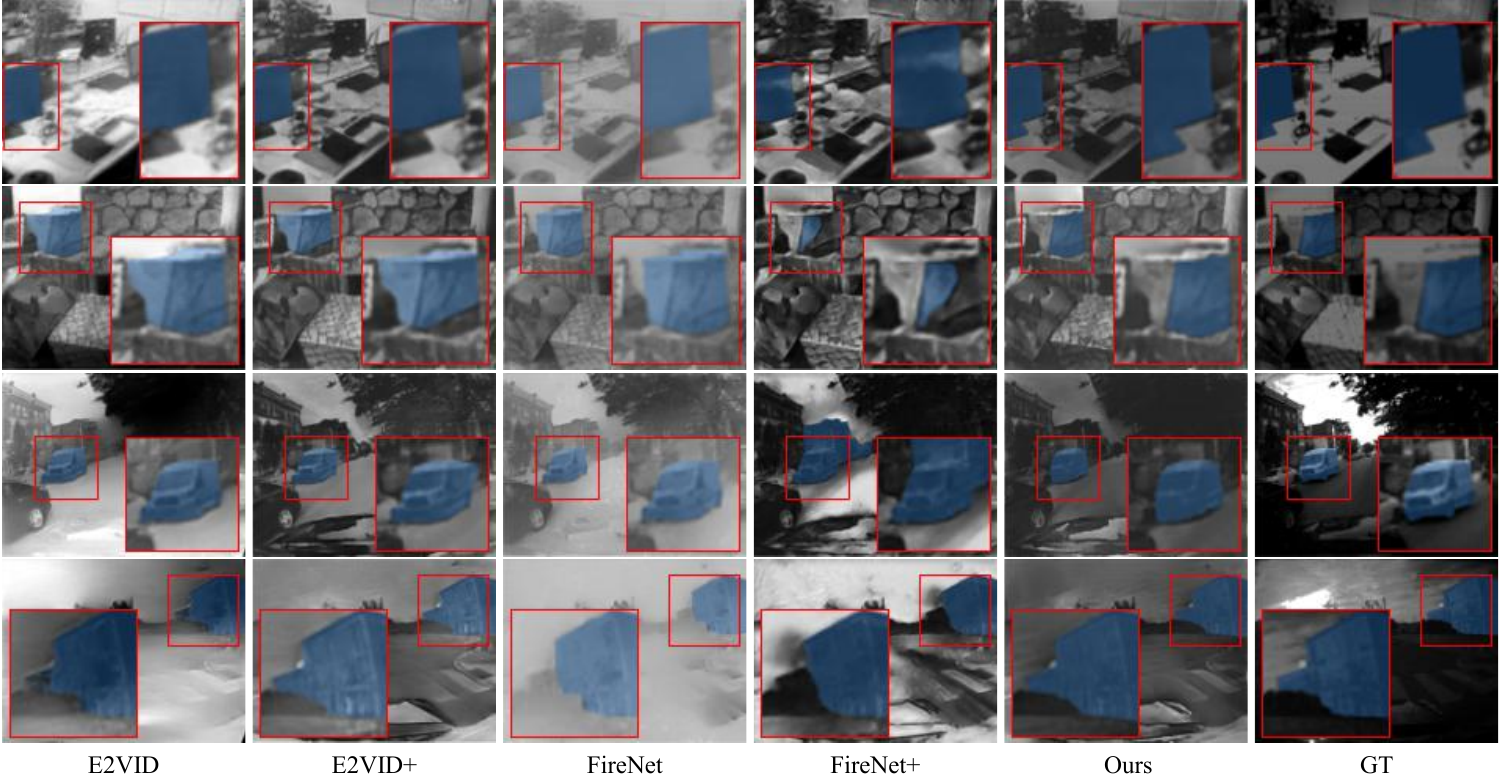} 
  \caption{Semantic mask extraction from reconstructed frames using SAM with point-based prompts.
Given the limited resolution and quality of event-based reconstructions, we avoid full-scene semantic segmentation and instead guide SAM using point prompts placed at visually or structurally salient regions.
Each row shows reconstructions from different methods, with identical prompt locations used for fair comparison.  The resulting semantic masks (overlaid in color) highlight key object boundaries and structures, providing localized but meaningful semantic cues.
  } 
  \label{fig:point_segment}
\end{figure*}

To evaluate the effectiveness of semantic knowledge in event-based reconstruction, we adopt a point prompting approach using the Segment Anything Model (SAM).
We select a single or a few prompt points at visually or structurally salient regions, typically centers of semantic objects.
These point prompts are used to generate localized semantic masks (Fig.~\ref{fig:point_segment}), allowing us to assess the semantic integrity of each reconstruction.
This strategy provides a controlled and robust way to extract semantic priors from low-quality reconstructions, enabling fair cross-method comparisons while highlighting the semantic alignment benefits introduced by our model.

\subsubsection{Qualitative Analysis on Semantic Feature Learning}
\begin{figure*}[h]
  \centering
  \includegraphics[width=0.95\linewidth]{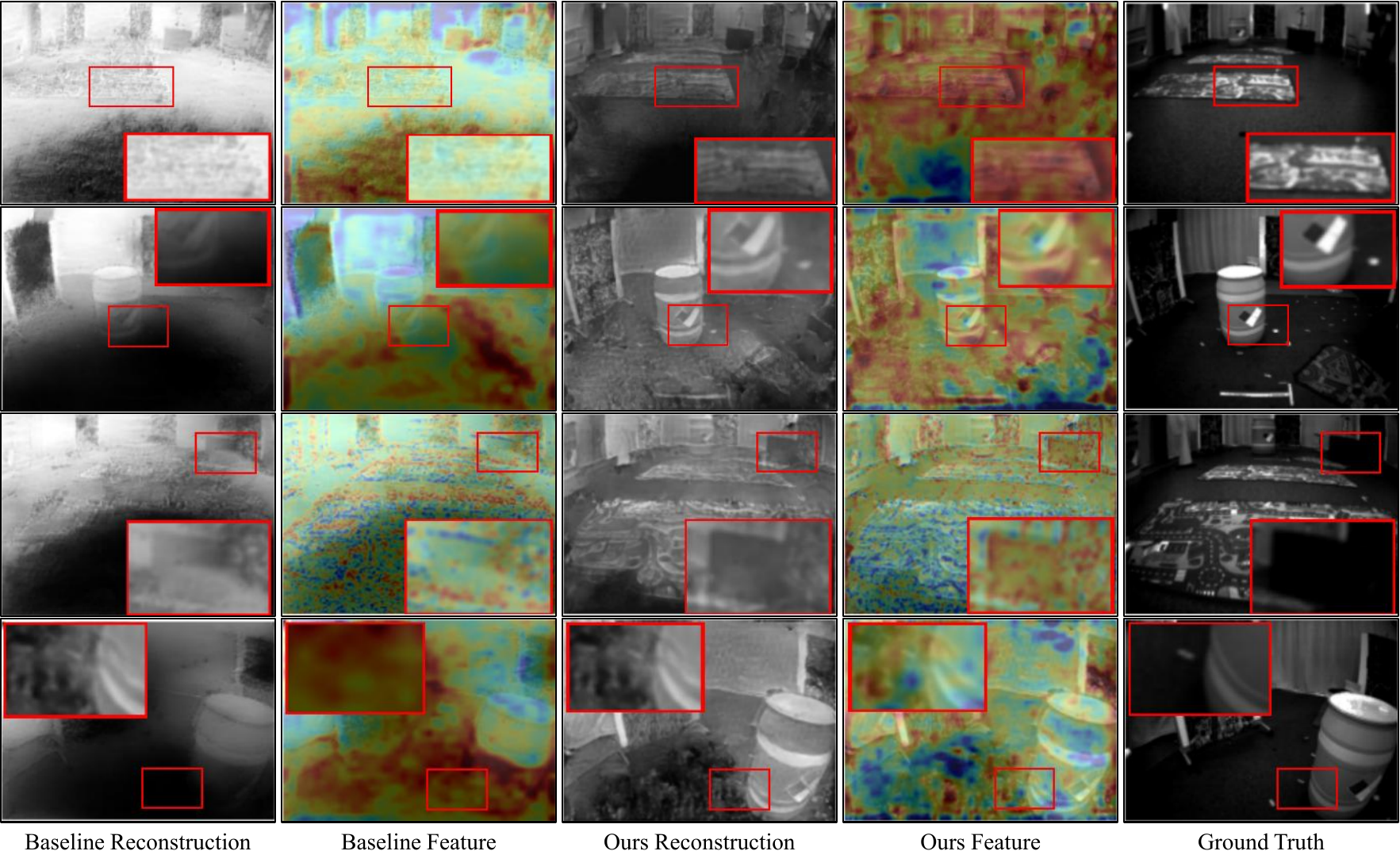} 
  \caption{Visualization of intermediate features and reconstruction results.
    Each row corresponds to one test scene. From left to right: reconstruction from baseline model, intermediate feature map of baseline, reconstruction from our method, our intermediate feature map, and ground truth. Our method produces more accurate reconstructions and learns features that better focus on semantically meaningful regions (e.g., objects, structural boundaries), as highlighted by the red boxes.} 
  \label{fig:feature_vis}
\end{figure*}

To better understand how our proposed semantic-guided learning module benefits the reconstruction process, we visualize the intermediate feature maps, $\mathcal{F}_e'$, and final reconstructions from our method, and compare the visualization of the intermediate feature maps $\mathcal{F}_e$, and final reconstructions from the baseline model in Fig.~\ref{fig:feature_vis}. We apply colormaps to visualize the feature activations, where warmer colors (e.g., red and yellow) indicate higher activation intensities, and cooler colors (e.g., blue) represent lower activations.

We observe that the baseline model often produces features with low activations in low-texture or background regions, resulting in blurry reconstructions, especially around key semantic structures (e.g., objects, furniture, human figures), as shown in the second column. This means the encoded feature $\mathcal{F}_e$ lacks semantic knowledge, and produces poor reconstruction results, especially for those semantic regions due to this (shown in the first column).
In contrast, through our proposed CFA and SFF, our method consistently produces features $\mathcal{F}_e'$ with rich semantic information, and learns to focus on semantically salient regions in event modality such as textures, objects, and surfaces that are crucial for visual understanding (shown in the fourth column). This targeted feature learning leads to notably improved reconstruction quality in those regions (shown in the third column). As shown in the red boxes, our method recovers finer details and structure in areas that the baseline struggles with, including sharper object contours and more accurate shading or texture. These results validate that our semantic-aware module not only guides the network toward learning perceptually important features but also enhances the reconstruction fidelity in those regions.

\subsection{Ablation Study} 
In this section, we provide a comprehensive quantitative evaluation of Semantic-E2VID, focusing on both the effectiveness of its designed components and its robustness under challenging conditions. We begin with detailed ablation studies to analyze the contribution of each core module—Semantic Abstraction, Semantic Feature Fusing, and Semantic-Aware Supervision-replacing or modifying them and measuring the resulting performance changes. We further extend the analysis to key hyperparameters, such as the loss weighting factor $\lambda$ and the top-$N$ semantic categories used in supervision, to verify that our method remains stable across a range of parameter settings.  Finally, we assess the robustness of Semantic-E2VID against real-world challenges, including varying event sparsity, different frame reconstruction intervals, and increased temporal irregularity. Results on the ECD and MVSEC datasets consistently show that Semantic-E2VID achieves the lowest MSE and highest SSIM compared to state-of-the-art baselines, highlighting its ability to produce accurate, stable, and semantically consistent reconstructions even under adverse conditions.

\begin{table*}[h]
\centering
\caption{Results on ECD, MVSEC, and HQF datasets with different ablation settings.}
\resizebox{\textwidth}{!}{
\begin{tabular}{c c c c c c c c c c c c c}
\toprule
\multirow{3}{*}{\textbf{Category}} & \multirow{3}{*}{\textbf{Setting}} 
& \multicolumn{3}{c}{\textbf{ECD}} 
& \multicolumn{3}{c}{\textbf{MVSEC}} 
& \multicolumn{3}{c}{\textbf{HQF}} \\
\cmidrule(lr){3-5} \cmidrule(lr){6-8} \cmidrule(lr){9-11}
& & MSE$\downarrow$ & SSIM$\uparrow$ & LPIPS$\downarrow$ 
  & MSE$\downarrow$ & SSIM$\uparrow$ & LPIPS$\downarrow$ 
  & MSE$\downarrow$ & SSIM$\uparrow$ & LPIPS$\downarrow$ \\
\midrule
\multicolumn{2}{c}{\textbf{Full Model}} 
& \textbf{0.014} & \textbf{0.594} & \textbf{0.208} 
& \textbf{0.067} & \textbf{0.329} & \textbf{0.483} 
& \textbf{0.030} & \textbf{0.577} & \textbf{0.219} \\
\midrule

\multirow{2}{*}{\textbf{Semantic Abstraction}} 
& Direct Distillation        
& 0.028 & 0.569 & 0.223 
& 0.087 & 0.302 & 0.501 
& 0.038 & 0.524 & 0.234 \\
& Effect                     
& {-0.014} & {-0.025} & {-0.015} 
& {-0.020} & {-0.027} & {-0.018} 
& {-0.008} & {-0.053} & {-0.015} \\
\midrule

\multirow{6}{*}{\textbf{Semantic Feature Fusing}} 
& Addition               
& 0.025 & 0.575 & 0.220 
& 0.078 & 0.305 & 0.500 
& 0.036 & 0.535 & 0.228 \\
& Effect                    
& {-0.011} & {-0.019} & {-0.012} 
& {-0.011} & {-0.024} & {-0.017} 
& {-0.006} & {-0.042} & {-0.009} \\

\cdashline{2-11}[2pt/1pt]

& Mean                
& 0.026 & 0.572 & 0.222 
& 0.079 & 0.303 & 0.502 
& 0.037 & 0.531 & 0.230 \\
& Effect                    
& {-0.012} & {-0.022} & {-0.014} 
& {-0.012} & {-0.026} & {-0.019} 
& {-0.007} & {-0.046} & {-0.011} \\

\cdashline{2-11}[2pt/1pt]

& Cross Attention        
& 0.020 & 0.584 & 0.216 
& 0.075 & 0.318 & 0.495 
& 0.036 & 0.545 & 0.225 \\
& Effect                    
& {-0.006} & {-0.010} & {-0.008} 
& {-0.008} & {-0.011} & {-0.012} 
& {-0.006} & {-0.032} & {-0.006} \\
\midrule

\multirow{2}{*}{\textbf{Semantic-Aware Supervision}} 
& Original LPIPS           
& 0.020 & 0.582 & 0.227 
& 0.072 & 0.313 & 0.505 
& 0.033 & 0.539 & 0.245 \\
& Effect                   
& {-0.006} & {-0.012} & {-0.019} 
& {-0.005} & {-0.016} & {-0.022} 
& {-0.003} & {-0.038} & {-0.026} \\
\bottomrule
\end{tabular}
}
\label{tab:Ablation}
\end{table*}

\subsubsection{Ablation Study on Designed Modules}
To comprehensively assess the effectiveness of each proposed component—namely the Cross-modal Feature Alignment (CFA) module, the Semantic-guided Feature Fusion (SFF) module, and the Semantic Perceptual E2V Supervision—we perform a series of ablation studies, as summarized in Table~\ref{tab:Ablation}. The Full Model integrates all three components.

\noindent\textbf{Semantic Abstraction.} 
This ablation investigates the impact of the CFA module. In the comparison setting, \textit{Direct Distillation} directly transfers the SAM-derived features $\mathcal{F}_{sam}$ into the event feature $\mathcal{F}_e$ output by the event encoder via relational distillation~\cite{park2019relational}, without leveraging the CFA module. The Semantic Perceptual E2V Supervision remains enabled. While this direct distillation superficially incorporates semantic knowledge, it introduces two critical issues. First, $\mathcal{F}_{sam}$ originates from the RGB frame domain, whereas $\mathcal{F}_e$ captures high-dimensional features from asynchronous event data. The naive injection of frame-based semantics into the event feature space causes modality misalignment, which misleads the event encoder during training. Consequently, the event encoder learns distorted or semantically inconsistent representations, harming the reconstruction fidelity.

This effect is evident in the quantitative results: on the ECD dataset, Direct Distillation lags behind our full method by $0.025$ in SSIM and $0.015$ in LPIPS, while on MVSEC the gaps are $0.027$ in SSIM and $0.020$ in MSE. These results validate the necessity of the proposed CFA module, which explicitly aligns cross-modal features before fusion, ensuring stable and semantically meaningful learning in the event domain.

\noindent\textbf{Semantic Feature Fusing.} This study evaluates different strategies for integrating semantically enriched event features into the original event representations, while disabling the SFF module. Three fusion methods are considered:  
\begin{itemize}
    \item \textit{Addition:} directly adds the CFA-learned features to the raw event features.
    \item \textit{Mean Fusion:} performs element-wise averaging between the learned features and raw features.
    \item \textit{Cross Attention:} applies a cross-attention mechanism to adaptively fuse the two feature streams.
\end{itemize}
All three configurations retain the CFA module and Semantic Perceptual E2V Supervision, thereby isolating the impact of the SFF module and fusion strategy.

While all variants attempt to inject semantic priors into the event representation, they fall short in various aspects. Both \textit{Addition} and \textit{Mean Fusion} operate in a spatially uniform and position-agnostic manner, ignoring the fact that semantic relevance varies significantly across spatial locations. This indiscriminate fusion dilutes important contextual cues and leads to feature ambiguity. On the other hand, \textit{Cross Attention} lacks precise spatial selectivity and incurs additional computational overhead without significantly improving alignment quality.

In contrast, our proposed SFF module employs a spatial attention mechanism and an adaptive fusion strategy to deeply integrate semantics at semantically relevant regions. By first generating a spatial attention map via convolutional transformations and global context modeling, SFF enables selective enhancement of event features where semantic signals are most influential. As a result, the fused representation $\mathcal{F}_{e}'$ retains high spatial fidelity and semantic consistency, leading to superior reconstruction.

Quantitative results further confirm this: our SFF outperforms Addition, Mean Fusion, and Cross Attention by margins of up to $0.032$ in SSIM on the HQF benchmark and $0.012$ in LPIPS on the MVSEC benchmark, demonstrating its effectiveness in leveraging semantics for high-fidelity video reconstruction.

\noindent\textbf{Semantic-Aware Supervision.} This ablation examines the role of the Semantic Perceptual E2V Supervision. Here, we replace the proposed semantically guided loss with a standard perceptual loss, while keeping both the CFA and SFF modules intact. This setup allows us to assess the supervision signal’s contribution to guiding semantic reconstruction.

Unlike conventional perceptual loss that compares low-level or middle-layer features in the frame space, our semantic perceptual supervision exploits the high-level semantics extracted from the Segment Anything Model (SAM). This semantically grounded loss encourages the network to align the reconstructed frames with not only the appearance but also the underlying object and region-level structure. In contrast, standard perceptual loss often focuses on texture similarity, which may overlook important semantic boundaries and mislead the decoder toward overly smoothed or locally inconsistent outputs.

Incorporating semantic-level supervision enables better regularization of the event-to-video translation, especially in cases where event streams are sparse or object shapes are ambiguous. Quantitatively, this substitution leads to noticeable drops of $0.019$, $0.022$, and $0.026$ in LPIPS across all three benchmarks, highlighting the critical role of semantic supervision in preserving structural fidelity and region-aware consistency during reconstruction.

\noindent\textbf{Conclusion.} These results collectively validate our architectural design: the CFA module effectively bridges the modality gap between events and semantics; the SFF module spatially refines this alignment; and the semantically guided supervision further reinforces consistency during reconstruction.

\subsubsection{Ablation Study on Hyperparameter}
Here, we further investigate the quantitative behavior of Semantic-E2VID through two key hyperparameter studies. We first analyze the sensitivity of the loss weighting factor $\lambda$, followed by the number of semantic masks $N$ derived from SAM to validate the stability of our design choices.

\noindent\textbf{Ablation on Hyperparameter in Loss Function.}
We conduct a robustness study on the hyper-parameter $\lambda$ in the final loss function (Eq.~\ref{eq: final_loss}). Table~\ref{tab:lambda_ablation} reports the reconstruction performance on the ECD dataset across a range of $\lambda$ values. Our chosen value of $\lambda = 1.8$ achieves the best performance in all three metrics—MSE, SSIM, and LPIPS. Notably, the performance variation is small across nearby values, demonstrating that our loss function is not overly sensitive to the exact choice of $\lambda$, which highlights the robustness of our formulation.

\begin{table}[h]
\centering
\scriptsize
\caption{Effect of hyper-parameter $\lambda$ in the final loss function $\mathcal{L}$. Results are reported on the ECD dataset.}
\label{tab:lambda_ablation}
\begin{tabular}{c|ccc}
\toprule
$\lambda$ & MSE $\downarrow$ & SSIM $\uparrow$ & LPIPS $\downarrow$ \\
\midrule
0.5  & 0.022 & 0.573 & 0.240 \\
1.0  & 0.019 & 0.581 & 0.227 \\
1.5  & 0.017 & 0.586 & 0.219 \\
\textbf{1.8}  & \textbf{0.014} & \textbf{0.594} & \textbf{0.208} \\
2.0  & 0.016 & 0.590 & 0.215 \\
\bottomrule
\end{tabular}
\end{table}

\noindent\textbf{Ablation on Hyperparameter for Number of Masks from SAM.}
To evaluate the impact of the number of semantic mask categories used in our semantic perceptual E2V supervision, we conduct ablation experiments by varying the top-$N$ non-background semantic classes extracted from SAM. Table~\ref{tab:topn_ablation} reports the performance with $N \in \{5, 7, 10, 15, 20\}$ on the ECD dataset. Using too few semantic masks (e.g., $N=5$) fails to capture rich object-level guidance, leading to degraded reconstruction. On the other hand, using too many masks (e.g., $N=20$) introduces noisy or redundant categories that may weaken the supervision signal. Our default choice $N=10$ achieves the best balance across all metrics (MSE, SSIM, and LPIPS), validating its effectiveness in guiding the E2V reconstruction with semantically meaningful perceptual cues.

\begin{table}[h]
\centering
\scriptsize
\caption{Ablation on the top-$N$ mask categories in semantic perceptual E2V supervision. Results are reported on the ECD dataset.}
\label{tab:topn_ablation}
\begin{tabular}{c|ccc}
\toprule
$N$ & MSE $\downarrow$ & SSIM $\uparrow$ & LPIPS $\downarrow$ \\
\midrule
5  & 0.021 & 0.575 & 0.246 \\
7  & 0.018 & 0.582 & 0.231 \\
\textbf{10} & \textbf{0.014} & \textbf{0.594} & \textbf{0.208} \\
15 & 0.016 & 0.589 & 0.217 \\
20 & 0.017 & 0.583 & 0.224 \\
\bottomrule
\end{tabular}
\end{table}

\subsubsection{Robustness Analysis}
\begin{figure*}[h]
  \centering
  \includegraphics[width=\linewidth]{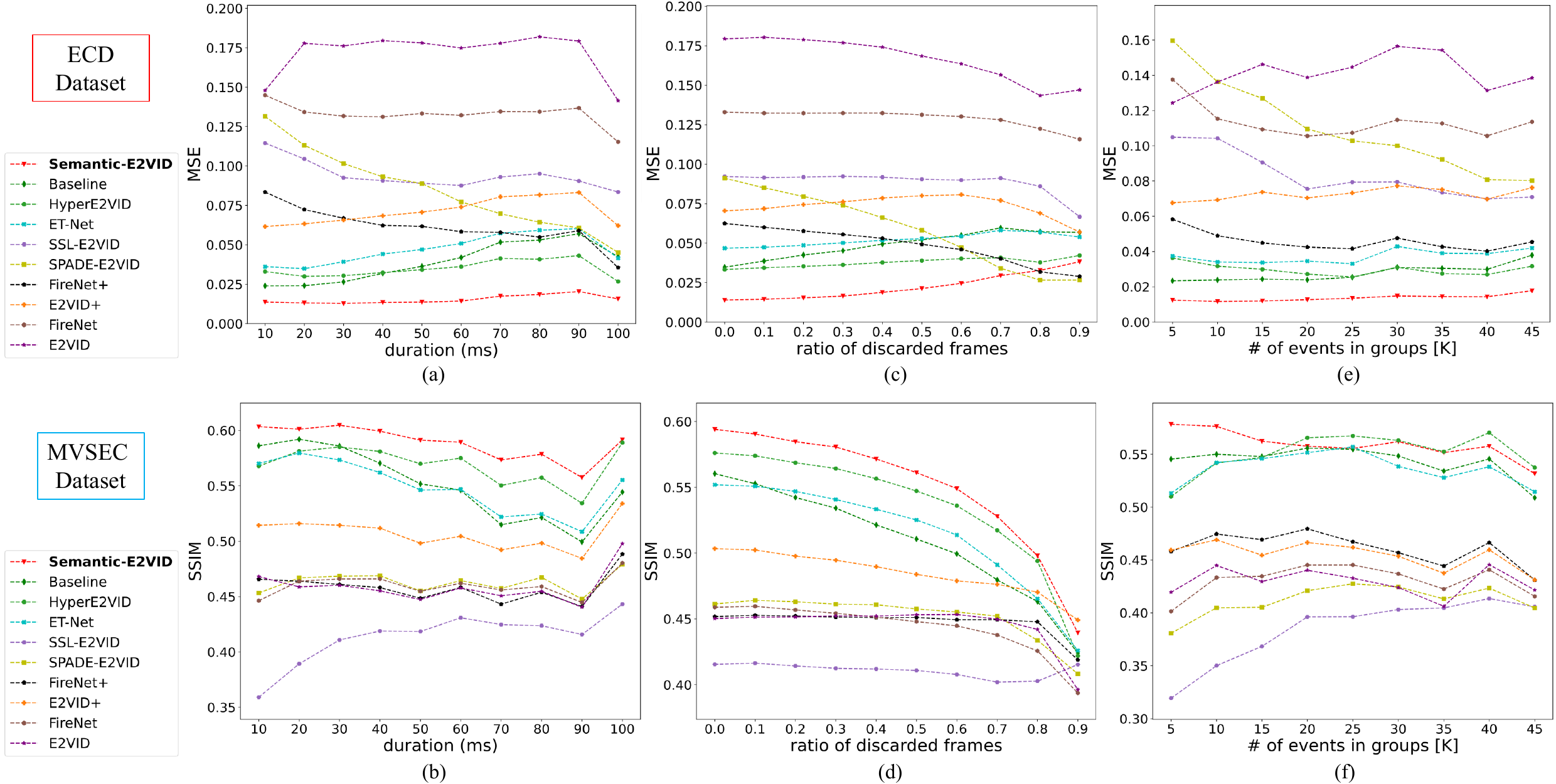} 
  \caption{We compare Semantic-E2VID with several state-of-the-art methods under three challenging conditions: (a, b) different frame reconstruction intervals (duration in milliseconds); (c, d) increasing temporal irregularity (ratio of discarded frames); and (e, f) varying levels of event tensor sparsity (number of events per voxel grid) on two critical metrics: MSE (a, c, e), and SSIM (b, d, f). Semantic-E2VID consistently achieves the lowest MSE and highest SSIM, exhibiting strong resilience to changes in event sparsity, reconstruction rate, and temporal inconsistency.}
  \label{fig:robust}
\end{figure*}

Reconstructing images from event data is a complex task influenced by multiple variables that impact method performance. To evaluate the robustness of our approach under varying conditions, we conducted robustness analysis by utilizing data sequences in the ECD and MVSEC datasets, using MSE and SSIM as the image quality metric. Specifically, we compare Semantic-E2VID with several state-of-the-art methods under three challenging conditions: varying levels of event tensor sparsity (number of events per voxel grid); different frame reconstruction intervals (duration in milliseconds); increasing temporal irregularity (ratio of discarded frames). Semantic-E2VID consistently achieves the lowest MSE and highest SSIM, exhibiting strong resilience to changes in event sparsity, reconstruction rate, and temporal inconsistency, shown in Fig. \ref{fig:robust}. Detailed aspects of robustness analysis are explained as follow:

\noindent\textbf{Reconstruction Rate.} To evaluate the robustness of our model to the impact of changing frame reconstruction rates, we adopt the fixed-duration grouping to control how many image frames are reconstructed per second. The grouping strategy ensures the duration of events in every voxel grid is fixed. In our experiment, the frame reconstruction interval is varied from $10$ milliseconds to 100 milliseconds in increments of $10$ milliseconds, resulting in ten distinct experimental settings. We still enforce that the corresponding reference image used to calculate the MSE metric is the nearest image to the reconstructed image in time, with a time tolerance of $1.0$ milliseconds. The results are shown in Fig. \ref{fig:robust} (a) and (b), Semantic-E2VID still performs better than other models and is minimally affected by changes in the reconstruction rate.

\noindent\textbf{Temporal Irregularity.} To evaluate the impact of reconstructing image frames at irregular intervals on our model, we utilize the between-frame event grouping method, where events between two consecutive image frames are aggregated together and used to generate one predicted image. We remove a certain percentage of frames in each data sequence randomly, with discarding ratios ranging from $0.0$ (standard case) to $0.9$, a total of $10$ runs. As shown in Fig. \ref{fig:robust} (c) and (d), our model performs well and the curve is relatively gentle. MSE rises slowly as the discarding ratio increases, higher than FireNet+ and SPADE-E2VID when the ratio gets too high.

\noindent\textbf{Event Tensor Sparsity.} To investigate the impact of event tensor sparsity on model performance, we systematically control the number of events aggregated into a single event voxel grid. Specifically, the number of events varied from $5000$ to $45000$ in increments of $5000$, nine experimental in total. In each inference step, the evaluated model receives a single voxel grid as input and produces a corresponding predicted image. The quality of the predicted image is computed against the temporally nearest ground truth reference image. A strict time tolerance of $1.0$ milliseconds is enforced between the predicted image and the reference image, ensuring minimal temporal discrepancy for accurate evaluation. The results shown in Fig. \ref{fig:robust} (e) and (f) demonstrate that our proposed model, Semantic-E2VID, outperforms these state-of-the-art methods. The curve is quite smooth, indicating that the change of tensor sparsity does not affect the performance of our model much.

\section{Limitations and Future Work}
\label{sec: limitation}
Although Semantic-E2VID demonstrates strong performance across multiple benchmarks, including grayscale frame, high dynamic range, and colorful reconstruction scenarios, some limitations remain. Specifically, the current semantic knowledge distillation is performed on a synthetic grayscale training dataset, which may not fully capture the complexity and diversity of real-world scenes. While the model performs well in general, there is still room for improvement in specific reconstruction tasks such as colorful RGB reconstruction, event-based nighttime reconstruction, and event-assisted overexposure adjustment.

In future work, we plan to explore more effective semantic distillation strategies using real-world datasets and investigate task-specific enhancements to improve performance in challenging scenarios. Additionally, extending the framework to support multi-task event-based scene understanding, such as joint reconstruction and semantic segmentation or depth estimation, holds promise for further expanding its applicability.


\section{Conclusion}
\label{sec:conclusion}
In this work, we revisited event-to-video reconstruction from a semantic perspective and identified semantic under-determinacy as a fundamental limitation of event-based sensing.
Unlike conventional E2V approaches that rely primarily on temporal aggregation and spatial feature modeling, we argued that faithful reconstruction from sparse event streams requires explicitly accounting for missing object-level structure and contextual information.
Based on this insight, we proposed \textit{Semantic-E2VID}, a semantic-enriched end-to-end E2V framework that reformulates reconstruction as a process of semantic abstraction, semantic feature fusion, and semantic-aware supervision.
By bridging event representations with semantic knowledge extracted from a pretrained foundation model in a representation-compatible manner, our approach enables event features to capture object-level structure and contextual cues that are missing in raw event streams.
Extensive experiments on six public benchmarks demonstrate that Semantic-E2VID consistently outperforms state-of-the-art methods, producing reconstructions with clearer object boundaries, stronger semantic consistency, and improved robustness under sparse or ambiguous event conditions.

\section*{Acknowledgments}
This work is supported by the Research Grants Council of Hong Kong (GRF 17201620) and the Theme-based Research Scheme (T45-701/22-R).

\bibliographystyle{elsarticle-num} 
\bibliography{reference}



\end{document}